\documentclass[12pt]{article}
\usepackage{amssymb}
\usepackage{latexsym}

\begin{document}

\title{A New Statistical Approach for Comparing  Algorithms for Lexicon Based Sentiment Analysis}

\author{Mateus Machado, Evandro Ruiz \& Kuruvilla Joseph Abraham\\
Department of Computation and Mathematics\\
University of S\~{a}o Paulo,
Ribeir\~{a}o Preto-SP, Brazil}
\maketitle 
\begin{abstract}
Lexicon based sentiment analysis usually relies on the identification of 
various words to which a numerical value corresponding to 
sentiment can be assigned.  In principle, classifiers can be obtained  from these algorithms 
 by comparison with human annotation, which is considered the gold 
standard. In practise 
this is difficult in languages such as Portuguese where there is a paucity of 
human annotated texts. 
Thus in order to compare algorithms, a next best step is to directly compare different algorithms with 
each other without referring to human annotation. 
In this paper we develop methods for a statistical comparison of algorithms 
which does not rely on human annotation or on known class labels. We will motivate the use of 
marginal homogeneity tests, as well as log linear models within the framework of 
maximum likelihood estimation 
We will also show how some uncertainties present in lexicon based sentiment analysis  
may be similar to those which occur in human annotated tweets. We will also show how the variability 
in the output of different algorithms is lexicon dependent, and quantify this variability in the output within the 
framework of log linear models. 
\end{abstract}

\section{Introduction}
\label{sec1}
The Web is universally recognised as a huge source of textual information in a wide variety of languages. 
The type of information contained may be expository in nature, factual, or may consist of opinions. 
Due to the sheer volume of information available on any given subject in any given language, the 
use of natural language processing tools to extract this information is a commonly used strategy. 
The extraction of opinions, evaluation and attitudes from free text, which goes by the name of 
sentiment analysis, is frequently implemented with the use of lexicons, or dictionaries, which can assign a 
polarity, positive, negative, or neutral to the adjectives which appear in the text. Alternatively, 
machine learning approaches which require manual classification can be used. 

Manual Classification, which is often considered the gold standard, requires 
a large corpus of texts with manual annotation. In languages such as portuguese
this is resource is not present. In a recent review on statistical methods for sentiment analysis, 
 \cite{stine}, these languages have been referered to as resource poor languages.  Even in languages such as English, in which there are 
many  annotated texts, problems can arise due the divergence of opinions between 
different annotators studying the same text \cite{Plostwitter}.  In \cite{Plostwitter}
a variety of tweets in different languages were annotated by multiple annotators,
including 9 separate annotators for English. Several of the English language 
tweets were analyzed by different annotators. In particular there are 
1144 tweets analyzed by both annotator 78 and annotator 80 (notation of \cite{Plostwitter}).
Each annotator assigns to each tweet one of three labels, positive, negative or neutral. 
The combined results of their analysis is shown in the contingency table below.
\begin{table}[!h]
\caption{\label{tab1}Comparison of Human Annotators}
\centering
\begin{tabular} {| c | c | c | c | }  \hline 
        &         N78 &  Ne78  &  P78 \\ \hline 
  N80  &       236   &    76   &     35 \\ \hline 
  Ne80 &         50   &   295   &    113 \\ \hline 
  P80 &       16      &  58   &    265 \\ \hline 
  \end{tabular} 
  \label{table:plospap} 
  \end{table} 
 
The presence of non zero off diagonal elements in the contingency table demonstrates a lack of 
agreement between these two annotators. Similar results can be obtained from 
other pairs of annotators. In addition to the lack of agreement between different 
annotators, the authors of  \cite{Plostwitter} show that the same annotator 
on occasion marks the same tweet with a different label. These results
complicate the use of manual annotation as a guideline for training a 
classifier. Given this variability in manual annotation, it is worthwhile to 
have a framework to directly compare algorithms without any reference to manual annotation. 
This is the main motivation for this paper. The statistical methodology which we will discuss later will be used to show (in a real dataset), that within the same lexicon 
the degree of concordance between algorithms is label dependent, and for the same labels the degree of 
concordance between algorithms is lexicon dependent. In particular, it will be shown that it is exceedingly 
unlikely that these differences can be due to chance variation. 

The question of 
comparing of machine learning classifiers for twitter feeds has been discussed earlier 
in \cite{tellez}, however in this paper unlike in \cite{tellez}, we do not make use 
of supervised classification and the methodology adopted is quite different. 

\section{Statistical Methodology} 
\label{section:statmeth} 
In order to motivate the choice of statistical methodology, we will consider 
the same set of texts analysed by two different algorithms. The output of each 
algorithm on any one of the texts is a label positive ($p$), neutral ($u$)  or 
negative ($n$). The output of both algorithms can be summarised 
in a contingency table of the form below. 
\begin{table}[!h]
\caption{\label{contin}Contingency Table for Comparison of Classifiers}
\centering 
\begin{tabular}{| c  | c  | c |} \hline
    $\;\;\;\;\;\;\; n_{nn}(n_{11}) \;\;\;\;\;\;\; $ & $\;\;\;\;\;\;\;  n_{np}(n_{12}) \;\;\;\;\;\;\;  $  & $\;\;\;\;\;\;\; n_{nu} (n_{13}) \;\;\;\;\;\;\; $ \\ \hline 
    $\;\;\;\;\;\;\; n_{pn}(n_{21}) \;\;\;\;\;\;\; $ & $\;\;\;\;\;\;\;  n_{pp}(n_{22}) \;\;\;\;\;\;\;  $  & $\;\;\;\;\;\;\; n_{pu} (n_{23}) \;\;\;\;\;\;\; $ \\ \hline 
   $\;\;\;\;\;\;\;  n_{un}(n_{31}) \;\;\;\;\;\;\; $& $ \;\;\;\;\;\;\;  n_{up}(n_{32}) \;\;\;\;\;\;\; $  & $\;\;\;\;\;\;\; n_{uu}(n_{33})\;\;\;\;\;\;\;  $ \\ \hline 
\end{tabular} 
\label{contin}
\end{table}

In Table 2,  $n_{pp}$ is the number of texts for which both algorithms
assign label $p$. $n_{pu}$ is the number of texts for which one algorithm
assigns label $p$ and the other label $n$. $n_{up}$ describes the same 
situation but with the labels of the algorithms switched. The values in parentheses 
$n_{11}$ {\em etc.} are the more common notation for the entries  this kind of table and 
will be used later on.  The presence of 
non vanishing off diagonal values indicates less than perfect agreement 
between the algorithms. However, it is still desirable to check the degree of concordance
between the methods, more specifically to see if the degree of concordance is less than 
would be expected by pure chance. The comparison of the performance between 
different evaluators arises in various contexts, for example when evaluating the degree of concordance 
between two radiologists each of which assign one of four qualitative labels to the same set 
of slides of ovarian tumors. Among the methods suggested in the statistical literature for 
the comparison between experts are marginal homogeneity testing and log linear modelling. 
We will describe each of  these methods in turn and later describe their use. The kappa 
of cohen \cite{kappa} , a single parameter which measures the degree of agreement 
between algorithms over and above that expected by pure 
chance has been frequently used for this purpose. However there are 
well known deficiencies associated with this measure {\it " In summarizing a contingency table by a single number 
the reduction of information can be severe". } \cite{Agresti}. Thus we will 
make use of other statistical methods as well. None of the methods we will describe assume knowledge of the true 
labels. Thus methods which rely on ranking classifier performance such as the 
 friedman test \cite{friedman}, whose use is suggested in \cite{Demesar} and 
\cite{BrownMues}, will not be considered in this paper. In addition, we will not make
use of methods in \cite{stine} which consider two  known class labels .

\subsection{Marginal Homogeneity Tests}
The first category of tests which will used are those which will help compare 
\begin{enumerate}
\item the difference in the proportion of texts labelled as $n$ by both algorithms,
\item the difference in the proportion of texts labelled as $p$ by both algorithms, and 
\item the difference in the proportion of texts labelled as $u$ by both algorithms. 
\end{enumerate} 
These are 
marginal homogeneity tests; for $3 \times 3$ contingency tables all three differences
can be tested simultaneously for statistical significance via the Stuart Maxwell Test \cite{SMtest1},\cite{SMtest2}.  
If the p-value obtained from this test is significant, then at least at one of the 
three differences is statistically significantly different from zero. The Stuart Maxwell test used here is a more general version  of the McNemar test 
whose use in the comparison of binary classifiers was suggested
in \cite{McNemar1}. 

\subsection{Log Linear Models}
In addition to marginal homogeneity tests we will also analyze 
contingency tables using log linear models. The use of log linear models
to study agreement among observers 
has been discussed before \cite{Agresti1992} , the same methodology
can be applied to discuss the agreement between algorithms. We will compare two models, 
the independence model and quasi independence models to test the concordance between 
algorithms to analyze polarity. The independence model may be written as 
\begin{equation}
 \log{n_{ij}} = \lambda + \lambda_{i}^{1} + \lambda_{j}^{2} 
 \label{equation:indep}
 \end{equation} 
 
  In this model the level of agreement 
 between algorithms which occurs is the level expected by 
 pure chance. 
 
 If we assume that the algorithms show a degree of 
 concordance which exceeds that expected by pure
 chance, then the simplest model we can consider is 
\begin{equation}
 \log{n_{ij}} = \lambda + \lambda_{i}^{1} + \lambda_{j}^{2} + \delta I(i=j) 
 \label{equation:quasi1}
 \end{equation} 

In equation \ref{equation:quasi1} the parameter $\delta$ enters 
only in the diagonal elements and describes the extent to which the values of diagonal 
elements differ from the values expected by pure chance.
However, Equation \ref{equation:quasi1} assigns the same parameter to describe 
the rate at which the expected values for the diagonal counts for all three labels 
differ from those expected by chance agreement 
between the two algorithms. In practise, the algorithms may have different 
degrees of concordance for different labels. To describe this situation 
we will use a specific type of  quasi independence model \cite{BFH} which can 
be written as 
\begin{equation}
 \log{n_{ij}} = \lambda + \lambda_{i}^{1} + \lambda_{j}^{2} + \delta_{i} I(i=j) 
 \label{equation:quasi}
 \end{equation} 

The $\delta_{i}$ parameter thus describes that contribution to the rate 
at which both algorithms assign label $i$, which cannot be explained by 
pure chance. There are three such 
parameters, and the diference between the sizes of these parameters
is an indication of the difference in the degree of concordance between 
the algorithms for each label separately. The use of quasi independence 
models to distinguish the degree of concordance and disagreement in human observers 
for each separate label has been suggested in \cite{Bergan}. 
 
 One question relevant to the analysis of concordance against disagreement 
which can be addressed through the 
use of log linear modelling is the following: if we have two 
observations which both algorithms classify as being different, for example both being either $n$ or $p$, 
what are the odds that both algorithms agree rather
than disagree on which observation is $n$ and which is $p$  ? Questions  
of this nature can be easily answered within the framework 
 of a quasi-independence model. If we denote by $\pi_{ij}$ the probability that for a given observation 
 one algorithm chooses label $i$ and the other label $j$, then as shown in 
 in \cite{Agresti}  the log of the odds can be estimated by 
 \begin{equation}  \log\left(\frac{\pi_{11}\pi_{22}}{\pi_{12}\pi_{21}}\right) =  \delta_{1} + \delta_{2}  \label{equation:odds} 
 \end{equation}  
 
 A large value for Equation \ref{equation:odds} would suggest a strong tendency to agree 
 rather than to disagree on what is $n$ and what is $p$. The value of the log odds is 0 if the the tendency to 
 agree is comparable with the tendency to disagree. 
 The same analysis can be repeated for all possible 
 combinations of labels, and for different lexicons leading to pairwise measures for agreement between 
 different labels. 
 
 We will also consider logarithm of the odds ratio which in the 
 framework of the quasi independence model is 
\begin{equation} \displaystyle \log\left(
\frac{\pi_{11}\pi_{33}}{\pi_{13}\pi_{31}} \right) -\log\left( \frac{\pi_{22}\pi_{33}}{\pi_{23}\pi_{32} }   \right)= \delta_{1} -\delta_{2}  \label{equation:oddsratio} 
\end{equation} 
This quantity may be used to see if the tendency to agree on what is $n$ and what is $u$, is comparable with the tendency to agree on what is $p$ and what is $u$. 

Due to finite sample sizes
however, model parameters cannot be estimated exactly. We will attempt
to estimate these parameters and confidence intervals 
using Maximum Likelihood Methods.  In addition, we will make use of 
the residual deviance to compare goodness of fit of 
the models discussed earlier  with 
the saturated model.  We will also make use of AIC scores to compare 
models and choose the most suitable model among those discussed. 
 
 \section{Data Sets and Algorithms} To illustrate the use of 
 these methods we will extend the results data presented in 
 \cite{propor}. In this paper, a well known Portuguese book 
 review corpus \cite{ReLi} was used analyzed  with different algorithms relying on 
 word polarities (which we denote by {\em Words}, adjective polarities (which we denote by {\em Adj}) among others. 
 Sentiment was assigned to the output by the use of  three separate lexicons, {\em OpLex} \cite{OpLex},
 {\em SentiLex} \cite{SentiLex}, and {\em LIWC} \cite{LIWC}. Additional Details about the Algorithms and lexicons can 
 be obtained from \cite{propor}. For 2233 texts from the \cite{ReLi} corpus, all the algorithms 
 discussed in \cite{propor}  and all dictionaries were used to assign a  sentiment  
 in the form of a  categorical label taking one of three possible values, 
 $p$, $u$, or $n$.  These outputs can be used 
 to perform various pairwise comparisons 
 between the algorithms. We will focus on the comparison between {\em Words} and {\em Adj}. 
 The comparison is performed using the methods 
 described in Section \ref{section:statmeth} using each 
  lexicon separately. All analyses are performed using the R programming language \cite{Rlang}, 
  and the Inter Rater Reliability (irr)\cite{irr}.  
   
  \section{Results}
  \subsection{Marginal Homogeneity Tests} 
 Using the {\em LIWC}  lexicon the contingency table obtained from the 
 cross comparison of the results from the {\em  Words} and the {\em Adj} 
 comparison is shown below 
 \begin{table}[!h] 
\caption{Words and Adj LIWC lexicon}
 \centering
    \begin{tabular}{|c  c | c | c | c |} \hline
 {\em Adj} $\rightarrow $  &   & $n$ & $p$ & $u$  \\ \hline 
{\em Words}  $\downarrow$  & $n$ & 55 & 4  & 97  \\ \hline 
 & $p$ & 49 & 637  & 1009 \\ \hline 
 & $u$ & 36 & 24  & 322 \\ \hline 
\end{tabular} 
\label{tab3}
\end{table} 
 In Table 3 the presence of numerous 
 non vanishing off diagonal elements 
 shows that the algorithms do not agree perfectly.  
 The tables for  {\em OpLex} and {\em SentiLex}
 share this feature and will not be shown here. 
 
In all three tables, the presence of relatively large diagonal elements 
suggests the degree of concordance between different labels is 
greater than would be expected by chance alone. In order to check 
this we check the values of the unweighted kappa of cohen 
for all three lexicons. The results are shown below

\begin{table}[!htb]
\caption{Values of Kappa} 
\centering
\begin{tabular}{|c | c | c | } \hline 
Lexicon & kappa & p-value  \\ \hline 
{\em LIWC} & 0.1731(0.1513-0.1949)  & $ < 10^{-10} $  \\ \hline  
{\em OpLex} & 0.5440  (0.5148-0.5731) & $ < 10^{-10} $ \\ \hline 
{\em SentiLex}  & 0.5820 (0.5537-0.6103)  & $ < 10^{-10} $ \\ \hline 
\end{tabular} 
\label{tabKappaValues}
\end{table} 

In order to test for differences in marginal frequencies we implement the Stuart Maxwell 
tests for all three lexicons, as mentioned earlier. The middle column shows that value 
of the test statistic, the final column the p-values.

\begin{table}[!h]
\caption{Stuart Maxwell Test Results } 
\centering
\begin{tabular}{|c | c | c | } \hline 
Lexicon & Test-Statistic & p-value \\ \hline 
{\em LIWC} & 1001  & $ < 10^{-15} $  \\ \hline  
{\em OpLex} & 165 & $ < 10^{-15} $ \\ \hline 
{\em SentiLex} & 158  & $ < 10^{-15} $ \\ \hline 
\end{tabular} 
\label{SMValues}
\end{table} 
 The p-value for 
 Stuart Maxwell test 
 is highly significant which is evidence against equal marginal frequencies of the labels ($n$,$p$, and $u$) 
 as obtained from the two algorithms. 
 
\subsection{Log Linear Models} 
To begin with, we analyze the independence model (equation \ref{equation:indep}),
for all three lexicons. Rather than specify the estimates and standard errors  for all 
parameters we show the deviance statistic and AIC score for all three lexicons. 
We also include the p-values obtained from a $\chi^{2}$ test with 4 
degrees of freedom.

\begin{table}[!h]
\centering
\caption{\label{tab4} Testing the Independence Model}
 \centering
    \begin{tabular}{| c | c  | c  | c | } \hline
 Dictionary  & AIC & Resid. Dev. & p-value  \\ \hline 
{\em LIWC}   & 439.73 & 373.36  & $ < 10^{-15} $   \\ \hline 
 {\em OpLex}  & 1433.2 &1363.4  & $ < 10^{-15} $ \\ \hline 
 {\em SentiLex}   & 1594.4 &  1524.4  & $ < 10^{-15} $  \\ \hline 
\end{tabular} 
\label{tabindep}
\end{table} 

From the p-values in this table we see that the independence model 
is unsatisfactory. In order to check whether a model of the form 
equation \ref{equation:quasi1} might  be a better fit we analyze the 
table of pearson residuals. 

\begin{table}[!htb]
\centering
\caption{\label{tab4} Pearson Residuals (Senti) }
 \centering
 \begin{tabular}{| c  | c  | c |} \hline
 27.86 & -15.32 &  -4.010 \\ \hline 
 -14.76 & 31.69 & -20.74 \\ \hline 
 -8.00 & -19.96 & 24.88\\ \hline 
\end{tabular} 
\label{PearsonSenti}
\end{table} 

In Table 7, we find that the pearson residuals have large positive values for the diagonal 
elements. This suggests that the independence model underestimates the degree of concordance between the 
algorithms. The tables for the pearson residuals for {\em LIWC} and {\em OpLex} share this feature and will not be 
shown here.  

The underestimation of the degree of concordance between the algorithms 
suggests  the use of an equation such as 
equation \ref{equation:quasi1} to obtain a better fit to the 
data. The results from equation \ref{equation:quasi1} are
presented for all three dictionaries. Rather than report all 
parameter estimates and standard errors the results for 
the $\delta$ parameter in equation \ref{equation:quasi1}
is shown along with the AIC score, the residual deviance and 
the corresponding p-value which is now obtained 
by comparison with the quantiles of $\chi^{2}_{3}$. 

\begin{table}[!h]
\centering
\caption{\label{tab4} Testing Single Diagonal Parameter}
 \centering
    \begin{tabular}{| c | c | c  | c  | c | } \hline
 Dictionary  & $\delta $ & AIC & Resid. Dev. & Resid p-value  \\ \hline 
{\em LIWC}   &  1.236 & 169.54 & 101.17  & $ < 10^{-15} $   \\ \hline 
{\em OpLex}  &  1.708 & 230.6 & 158.53   & $ < 10^{-15} $ \\ \hline 
{\em SentiLex}   & 1.828 & 187.78  &  115.86  & $ < 10^{-15} $  \\ \hline 
\end{tabular} 
\label{tabsquasi}
\end{table} 

For all three dictionaries, the p-values associated with the $\delta$ 
parameter are small $ ( < 10^{-10}$) which is evidence against 
the value zero, the null hypothesis.  However, 
for all three dictionaries the p-value associated with the residual deviance is significant,
this model does not provide an adequate fit to the 
saturated model. The next step is to implement the quasi independence 
model (equation \ref{equation:quasi}) in the hope of obtaining a better 
fit. The AIC and Deviance Statistics for the Quasi Independence Model 
are shown below. 
\begin{table}[!h]
\centering
\caption{\label{tab4} AIC and Deviance Statistic Quasi Independence}
 \centering
  \begin{tabular}{| c | c  | c | c |} \hline
  Dictionary & AIC & Resid. Dev.  & Resid. p-value \\ \hline 
  {\em LIWC} & 72.53 & 0.1600 & 0.6891 \\ \hline 
  {\em OpLex}  & 75.83 & $4.038 \times 10^{-3}$  & 0.9493 \\ \hline 
  {\em SentiLex} & 77.56 & 1.636 & 0.2008 \\ \hline 
  \end{tabular}
  \label{tabquasiAIC}
  \end{table} 
  
Rather than report the estimates and standard errors for all parameters 
we will report the sample estimates of the parameters $\delta_{1}, \delta_{2}$  and 
$\delta_{3}$ (corresponding to labels $n$, $p$ and $u$) 
along with  their 95\% confidence intervals obtained from the profile likelihood. 

\begin{table}[!h]
\centering
\caption{\label{tab4} Testing Quasi Independence}
 \centering
    \begin{tabular}{|c| c | c | c  | c |} \hline
Lexicon &  Parameter  & Estimate & 95\% CI  & p-value  \\ \hline 
{\em SentiLex} &  $\delta_{1}$   &  2.9741 &  (2.586,3.386)  &$ < 10^{-9} $   \\ \hline 
{\em SentiLex} & $\delta_{2}$  &  3.1175 & (2.766,3.492) &  $ < 10^{-9} $ \\ \hline 
 {\em SentiLex} & $\delta_{3}$  & 0.06139 & (-0.3227,0.4268)  &  0.7475  \\ \hline 
{\em OpLex} & $\delta_{1}$   &  3.005 &  (2.610,3.426)  &$ < 10^{-9} $   \\ \hline 
{\em OpLex} & $\delta_{2}$  &  3.269 & (2.905,3.659) &  $ < 10^{-9} $ \\ \hline 
{\em OpLex} &  $\delta_{3}$  &  -0.4004 & (-0.800,-0.02277)  &  0.0429  \\ \hline 
{\em LIWC} & $\delta_{1}$   &  2.4371 &  (2.012,2.865)  &$ < 10^{-9} $   \\ \hline 
{\em LIWC} & $\delta_{2}$  &  2.9125 & (2.405,3.449) &  $ < 10^{-9} $ \\ \hline 
{\em LIWC} &  $\delta_{3}$  &  -0.8019 & (-1.224,-0.3776)  &  0.0002  \\ \hline 
\end{tabular} 
\label{tabsquasiliwc}
\end{table} 

Next we will present for each dictionary, the results for the 
log of the odds defined in equation ~\ref{equation:odds} . We present 
not only the sample estimates but also the 95\% Normal confidence intervals. 

\begin{table}[!h]
\centering
\caption{\label{tab4} Log Odds}
 \centering
    \begin{tabular}{|c | c | c | c |} \hline
Lexicon &  Label Pairs  & Estimate & 95\% CI   \\ \hline 
{\em LIWC} & $n$ and $p$  &  5.350 &  (4.606,6.093)    \\ \hline 
{\em LIWC} & $n$ and $u$  &  1.635 & (1.159,2.111)   \\ \hline 
{\em LIWC} &  $p$ and $u$  &  2.111 & (1.707,2.514)  \\ \hline 
{\em SentiLex} & $n$ and $p$  &  6.092 &  (5.456,6.727)    \\ \hline 
{\em SentiLex} & $n$ and $u$  &  3.036 & (2.681,3.390)   \\ \hline 
{\em SentiLex} & $p$ and $u$  &  3.179 & (2.915,3.443)  \\ \hline 
{\em OpLex} & $n$ and $p$  &  6.2744&  (5.610,6.939)    \\ \hline 
{\em OpLex} & $n$ and $u$  &  2.605 & (2.246,2.964)   \\ \hline 
{\em OpLex} &  $p$ and $u$  &  2.869 & (2.614,3.124)  \\ \hline 
\end{tabular} 
\end{table} 

Finally we show for each lexicon  the log of the Odds Ratio as defined by 
equation ~\ref{equation:oddsratio}. Once again, we use 95\% Normal 
Confidence Intervals. 

\begin{table}[!h]
\centering
\caption{\label{tab4} Log Odds Ratio}
 \centering
    \begin{tabular}{| c | c | c |} \hline
Lexicon &  Estimate & 95\% CI   \\ \hline 
{\em LIWC} & -0.4754 & (-1.0683,0.1175 )\\ \hline 
{\em SentiLex} & -0.1434 &  (-0.5646,0.2777)  \\ \hline 
{\em OpLex} & -0.2637 & (-0.6800,0.1525)  \\ \hline 
\end{tabular}
\end{table} 

\section{Discussion}
A variety of different conclusions can be drawn from the results presented 
in the various tables. Tables 4, and 5 show that independent of the 
dictionary used the algorithms show substantial disagreement. The 
values for the kappa of cohen suggest are similar for the 
{\em OpLex}  and the {\em SentiLex} with the  results from the {\em LIWC}  dictionary
quite distinct from the other two. The results of the Stuart Maxwell Test 
again are similar for the {\em OpLex} and {\em SentiLex}, but the results from 
{\em LIWC} are again quite distinct. This suggests that {\em LIWC} is 
somewhat different from the other two lexicons. 

However, the analysis of marginal homogeneity serves only to compare 
marginal frequencies. In order to check the rate at which the same text is given 
the same label by different algorithms, the results from the log linear analysis
are required. 

The results of the analysis of the independence model, Equation \ref{equation:indep}, shown in (Table 6) suggest that 
independent of the dictionary used, the two algorithms agree at a 
rate higher than expected by chance.  However, the low p-values obtained from the deviance 
residual are a motivation to consider more complex models. This 
conclusion holds independent of the dictionary used. Table 7 suggests that the 
diagonal elements are underestimated in the independence model. 

The first model used to address the inadequacies of the independence 
model is that described in Equation \ref{equation:quasi1}. The
results shown in Table 8 confirm the suggestion from the 
Table 6 about the inadequacy of the independence model. More 
specifically, an estimated value for the $\lambda$ parameter,  
statistically significantly different from 0, 
suggests that, independent of the dictionary, the observed level 
of agreement between the two algorithms is different from that 
expected by chance. This is compatible with the 
positive diagonal residuals in Table 7. 

Another measure for the improvement of this model 
in comparison with the independence model is the 
lower value for the AIC compared with the 
independence model, independent of the dictionary used.
Despite this improvement, the low p-value associated with residual
deviance suggests strongly that this model provides 
a poor approximation to the saturated model. This justifies the 
consideration of the quasi-independence model, Equation ~\ref{equation:quasi}. 

Proceeding to the results of the quasi independence model, Equation \ref{equation:quasi}, shown in Table 9, we find that the 
AIC scores are lower than for any other model considered earlier 
and the p-value associated with the residual deviance is sufficiently 
large to be considered a reasonable substitute for the saturated model. Since the quasi independence 
model is the only model considered which shows no evidence for overdispersion, we will consider 
only the quasi-independence model from now on. 

The values of the three $\delta$ parameters (Table 10) 
show different patterns for the three dictionaries. In the case of {\em LIWC} they 
are consistently lower than for {\em OpLex} and {\em SentiLex}. In common with 
Tables 4, 5 and 6 {\em LIWC} seems to be different. In the case of {\em SentiLex} it is possible 
to consider a simpler model in which the parameter $\delta_{3}$ is dropped from the model. 
However, we will retain the parameter $\delta_{3}$ in all lexicons, this does not introduce any changes to the 
conclusions drawn from the results of Tables 11 and 12. 

In Table 11 the estimated value of the Log Odds for the $n,p$ combination is larger than for the 
other two combinations. Along with the confidence intervals, this can be interpreted to mean that for all three lexicons 
the separation between $n$ and $p$ is much more distinct than between $n$ and $u$ and 
$p$ and $u$. This is consistent with the results obtained in \cite{Plostwitter} for the comparison 
of human annotators which were obtained using different statistical tests. This suggests that 
the relative difficulty of human annotators in distinguishing $n$ and $u$ and $p$ and $u$, compared
with $n$ and $p$, may get carried over to lexicon based polarity analysis. In Table 12, independent of 
lexicon, the confidence intervals for all three logs of the odds ratio includes the value zero. This can be interpreted
to mean that within a given lexicon, the tendency to agree on what is $n$ and what is $u$ is comparable with the tendency to 
agree on what is $p$ and what is $u$. However the tendency to agree on what is $n$ and what is $p$ is very much larger. 
In other words $u$ can be considered lie midway between $n$ and $p$. This seems to be compatible with the results of 
\cite{Plostwitter}. As an additional check on the consistency between the results shown here and those in \cite{Plostwitter}
Table 1 was analyzed using Equations \ref{equation:quasi},\ref{equation:odds} \& \ref{odds:ratio}. The results obtained show
the same trends as Tables 11 and 12. 

As in the case of the earlier analyses, there are perceptible lexicon dependencies in Table 11. 
The results for {\em OpLex} and {\em SentiLex} are more 
similar to each other than the results for {\em LIWC}, similar to the analysis of Tables 4 and Table 5. The 95\% confidence 
intervals in Table 11 for the {\em LIWC} $(n,u)$ Log Odds and $(p,u)$ Log Odds do not overlap with the confidence intervals 
of the $(n,u)$ Log Odds and $(p,u)$ Log Odds  for the other two lexicons. The non overlap of the confidence intervals, along with 
the smaller values of the Log Odds for {\em LIWC} 
compared with corresponding values for the other two lexicons, suggest that 
for the label pairs $(n,u)$ and $(p,u)$, the level of concordance between {\em Words} and {\em Adj} is 
statistically significantly less in {\em LIWC} than in {\em OpLex} or {\em SentiLex}. In particular, given the 
lack of overlap in the confidence intervals, this effect is exceeding unlikely to be due to sampling variation

The dictionary dependence of lexicon based sentiment analysis has been 
commented on qualitatively earlier \cite{souza} in the context of a single algorithm in conjunction with 
various lexicons. The analysis presented in this paper deals with the 
difference between the results of two algorithms 
studied across multiple lexicons and is more quantitative. Furthermore, unlike in \cite{souza} the methods presented here  do not 
make use of known class labels. From our analysis,
it appears that within a given lexicon the results of different algorithms show statistically significant differences, 
and furthermore these differences themselves can be lexicon dependent. These results were 
obtained through the use of marginal homogeneity tests and 
log linear modelling, which illustrates the advantage of these methods. 

To summarise, we have employed a statistical technique well known in the 
comparison of human evaluators to compare algorithms for lexicon based 
sentiment analysis. We find that independent of the lexicon used, the 
concordance between different algorithms on the pair of labels $n$ and $p$ appears to be 
much better than the concordance on the pairs $n$ and $u$ or  $p$ and $u$. 
This is consistent with the results obtained from a comparison of human evaluators
of twitter feeds. We have also shown that the degree of concordance between 
different algorithms is lexicon dependent. This conclusion has been backed up 
by extensive statistical analysis, including numerous p-values and confidence intervals
and extends previously published work on lexicon dependence.
One open question in this context is how much of this variability in the degree of 
concordance is merely due to the variability in the lexicons, and how much is 
due to the variability in the algorithms. This question requires further investigation. 

Quite apart from the data addressed here, the statistical methodology proposed 
has many other potential applications and is well suited to shed light on the 
details on the differences and similarities in the output of different algorithms 
whenever the output is in the form of a set of discrete labels which are mutually 
exclusive and complete. Furthermore, 
the statistical methodology discussed can also be used to analyze confusion
matrices obtained from the comparison of true and predicted labels. 

One limitation of the approach discussed in this paper is that Table 3 
cannot have too many vanishing entries, if there are many vanishing 
entries the maximum likelihood estimators may not exist. In such instances 
the methods discussed by \cite{Rapallo} may be better suited for the kind 
analysis discussed in this paper. 

\section{Acknowledgments}
The results obtained here were obtained as part of a Project Funded by the Pr\'{o} Retoria de Pesquisa of the 
University of S\~{a}o Paulo, Call No. 668/2018, Project No. 18.1.1719.59.8 

\bibliographystyle{plain}
\bibliography{myrefs_short}

\end{document}